# Chapter 8 The interface of intonation and lexical tone: Boundary phenomena in Mandarin varieties

**Cong Zhang & Yiya Chen**

**Abstract:**

This chapter explores the intricate interplay between intonation and tone in Mandarin Chinese varieties, focusing on f0, the primary acoustic cue for both intonation and tone. The main empirical base is intonation boundary phenomena, where intonation and tone intersect and influence each other in conveying a range of sentence-level linguistic functions – such as question vs. statement – and a rich array of speakers' attitudinal information. Theoretical models and emerging techniques are also discussed to account for the observed interactions of tonal aspects and boundary phenomena to convey multiple levels of communicative meanings.

## 8.1 Introduction

The interplay between intonation and lexical tone stands as a captivating domain of inquiry in both sound structures (i.e., phonetics and phonology) and sound-meaning mappings. In a lexical tone language, both intonation and lexical tone utilize similar acoustic cues to form the melodic components of speech. Intonation encodes structures and functions at the utterance level with pitch variations, while lexical tone specifies lower lexical-level pitch variations on individual tone-bearing units (such as syllable and word). Distinguishing the functions of these similar acoustic cues (primarily f0), i.e., teasing apart tone and intonation, is known to be complicated due to their intricate interactions (see also the Franconian dialects in Chapter 6).

Over the last decades, increasingly more effort has been made to understand the complex relationship between tone and intonation. This chapter explores such interactions, focusing on Mandarin Chinese varieties (see Chen 2022a for a broader review of tone-intonation interaction in Sinitic varieties). Mandarin Chinese, known for its lexical tonal system and intonational nuances, provides a fertile ground for investigating these dynamic interactions.

Through exploring empirical research, theoretical frameworks, and acoustic analysis, this chapter aims to illuminate how intonation and lexical tone intersect, influence each other, and jointly contribute to the prosodic landscape of Mandarin Chinese. In doing so, we seek to examine the mechanisms underlying the production, perception, and interpretation of intonational patterns, via the lens of the distinctive tonal structures of Mandarin Chinese varieties.

Mandarin Chinese is a contour-tone language (Yip 1989), with f0 serving as the primary acoustic correlate that captures the pitch patterns associated with the contour tones. Furthermore, f0 is also an important parameter for intonation, encompassing the melodic pitch patterns in connected speech. By manipulating f0, speakers can express emotions, attitudes, sentence types, and emphasis. Note that we do recognize that many other acoustic measurements, including amplitude, voice quality, duration, and spectral measures, contribute to signal tone and intonation. This chapter, however, limits its attention to f0, and only briefly discusses other cues, due to the primacy of f0 for lexical tone. Additionally, our primary focus will be on boundary phenomena due to space limitations.

The rest of this chapter will begin with an introduction to the tonal aspects of Mandarin Chinese varieties, followed by an overview of boundary phenomena across different sentence types. We will then zoom into specific types of boundary phenomena observed in corpora, highlighting the key topic of how intonation interacts with lexical tones. Subsequently, we will discuss classic theoretical models and emerging techniques to enhance our modelling of intonation. By combining empirical findings with theoretical perspectives, we aim to shed light on the multifaceted nature of the intonation-tone interface, enriching the literature on sentence- and word- level prosody interaction in world languages by showcasing how tone and intonation can interact in different ways.

## 8.2 Mandarin Chinese varieties and cross-dialect tonal variations

We start with some background about Mandarin Chinese. Mandarin is one of the ten dialectal groups that belong to the Sinitic languages spoken in China, which include Mandarin (官话), Jin (晋), Wu (吴), Hui (徽), Gan (赣), Xiang (湘), Min (闽), Ke (客, Hakka), Yue (粤, Cantonese), and Ping (平) (Wurm et al., 1987). Mandarin can be further divided into eight subgroups of dialects (*pian*, 片). Within each subgroup, there are various dialectal clusters (*xiaopian*, 小片) which consist of different local dialects (*dian*, 点).

The official language of mainland China is *Putonghua* (普通话, “common speech”), commonly known as “Standard Mandarin” or “Standard Chinese”. This standardized lingua franca is primarily based on the vocabulary of northern Mandarin and follows the grammar of written vernacular Chinese. In Taiwan, a parallel variety is called *Guoyu* (国语, “national language”) or Taiwan Mandarin, while in Singapore, it is referred to as *Huayu* (华语, “Chinese language”). All three varieties share significant similarities but remain distinctive.

Many Mandarin varieties are mutually intelligible, especially between those within the same subgroup, but they differ significantly in lexical tones. Standard Mandarin includes four lexical tones and a neutral tone (Chen and Xu 2006; see Chen 2013 for a review of neutral tones in different Mandarin dialects). Table 8.1 illustrates their pitch contour shapes, tone numbers, and tonal compositions. The tone numbers, a.k.a. “Chao letters”, were developed by Chao (1930). This system divides the whole tonal space into five equal pitch levels,[1] with 1 being the lowest and 5 the highest. It has been the convention among Sinologists to represent tones with tone numbers. Within the generative linguistic framework, lexical tones are analyzed to be composed of discrete High (henceforth H) and Low (henceforth L) tones (e.g. Yip 1980). In addition, pitch register has been proposed to distinguish contrastive lexical tones in some dialects, as shown in Table 8.2, where the register difference is labelled using lowercase l(ow) and h(igh).

**Table 8.1: Lexical tones of Standard Mandarin**

| **Tone** | **Contour shape** | **Tone number** | **Tone composition** |
|---|---|---|---|
| **Tone 1** | Level | 55 | HH |
| **Tone 2** | Rising | 35 | LH |
| **Tone 3** | Dipping/Low | 214 | LLH |
| **Tone 4** | Falling | 51 | HL |

<insert Table 8.1 here>

To illustrate the tonal differences between closely related Mandarin varieties, Table 8.2 lists four other Mandarin varieties, including Tianjin Mandarin (Li et al. 2017), Kaifeng Mandarin (Wang et al. 2020), Xi’an Mandarin (M. Liu et al. 2020), and Zhushan Mandarin (Chen and Guo 2022). Take the character, 花 /hua/ ‘flower’, as an example. It carries a Tone 1 in Standard Mandarin. Despite the fact that this character carries the same tone category across the four varieties, they have different f0 contours and consequently, tone numbers: 55 in Standard Mandarin, 31 in Tianjin Mandarin, 24 in Kaifeng Mandarin, 21 in Xi’an Mandarin, and 324 in Zhushan Mandarin. Across Mandarin dialects, we observe significant variations in the pitch patterns of the same lexical tone category.

[1] Researchers define these levels differently. See Xu & Zhang (2024) for a meta-analysis review.

**Table 8.2: Lexical tones of four Mandarin varieties**

| | Contour shape | Tone number | Tone composition |
|---|---|---|---|
| **Tianjin Mandarin** | | | |
| **Tone 1** | Low-falling | 31 | l-HL |
| **Tone 2** | High-rising | 45 | h-LH |
| **Tone 3** | Low-dipping/low-rising | 213 | l-LH |
| **Tone 4** | High-falling | 53 | h-HL |
| **Kaifeng Mandarin** | | | |
| **Tone 1** | Rising | 24 | LH |
| **Tone 2** | Falling | 41 | HL |
| **Tone 3** | High level | 55 | H |
| **Tone 4** | Falling/Dipping/Low | 31/312 | L |
| **Xi'an Mandarin** | | | |
| **Tone 1** | Low-falling | 21 | l-HL |
| **Tone 2** | Mid-rising | 24 | m-LH |
| **Tone 3** | High-falling | 53 | h-HL |
| **Tone 4** | High-level | 55 | HH |
| **Zhushan Mandarin** | | | |
| **Tone 1** | Low-register rising | 324 | l-LH |
| **Tone 2** | High | 54 | H |
| **Tone 3** | High-register rising | 435 | h-LH |
| **Tone 4** | Falling | 51 | HL |

<insert Table 8.2 here>

### 8.3 Boundary phenomena

The main empirical focus of this chapter is the intonational phenomenon at or near utterance-level boundaries. We will begin by reviewing intonational boundaries in statements and questions, which have been the primary topic of investigation in most studies on Mandarin intonation in the literature have (but see Zhang 2018a; 2018b for chanted call tunes in Tianjin Mandarin).

#### 8.3.1 Boundary of statements

A predominant phenomenon surrounding statements in literature on Mandarin Chinese is the presence or absence of declination at the end of an utterance. Declination refers to the tendency of a gradual f0 decline across an utterance (Cohen and 't Hart 1965; Ladd 1984) and has been observed in many languages (e.g., Liberman and Pierrehumbert 1984 for English; Laniran and Clements 2003 for Yoruba; Prieto et al. 1996 for Spanish; and Scholz and Chen 2014 for Wenzhou Wu Chinese). Declination has been argued to be planned (e.g., Fuchs, et al., 2013; Scholz & Chen, 2014; Kim & Tilsen, 2024) despite that it is mainly a global f0 trend rather than a boundary-specific phenomenon (c.f. final lowering). Its relevance with boundary events makes it pertinent to our discussion of the Mandarin Chinese statement intonation.

Researchers differ as to whether Mandarin Chinese has declination. Tseng (1981) studied spontaneous and read speech in Taiwan Mandarin and showed that only 20% of the utterances in spontaneous speech showed a declination effect, which led her to argue that the declination effect is not common in Mandarin. Xu (1999) took a step further and examined declination using declarative utterances from eight native speakers of Standard Mandarin. He concluded that the downtrend in Mandarin is mainly determined by lexical tone identity. The overall f0 downward trend with exclusively high tones was only 0.9 Hz. As the number of L tones in an utterance increased, a more pronounced declination was observed, with a maximal of 39.5 Hz with the LRL (Low-Rise-Low) tone sequence. This led Xu to propose the presence of downstep triggered by low tone targets in Mandarin Chinese.

Evidence for declination in Mandarin, however, has also been reported. Shih (1997; 2000) investigated lab speech from four Mandarin speakers, two speaking Standard Mandarin and two speaking Taiwan Mandarin. They produced two utterances with a narrow focus. A global declination effect (of approximately 50 Hz as shown in the figures) was observed with a greater magnitude at the beginning. Note that there were contrasts between the initial subject part of the utterances, which might have led to an exaggerated declination effect due to on-focus f0 raising and/or post-focus f0 lowering. In another study, Wang and Lin (2003) examined a natural-occurring Standard Mandarin telephone conversation corpus and reported declination in a number of sentence types, including *wh*-questions, polar questions, and statements. Yuan and Liberman (2014) also observed f0 declination in spontaneous speech using a Mandarin broadcast news speech corpus, which they argued to be linguistically controlled rather than a by-product of the physics and physiology of talking.

The discrepancies reported in the literature may be caused by a range of factors. In addition to different experiment designs and speech materials reported above, the difference may also arise from different data modelling techniques. For instance, both Yuan and Liberman (2014) and Shih (2000) modelled the f0 curves using different regression models, while Xu compared the first and the last tones directly within an utterance. Similarly, concerning the more localized final lowering phenomenon, Shih (2000) compared different utterances with two conditions (with and without the final neutral-toned particle) and reported no final lowering. Yuan and Liberman (2014), however, modelled the topline (the line connecting local f0 peaks) and found a sharper fall towards the end of the same utterance, concluding the existence of final lowering. These methodological differences might have played an important part in the different conclusions drawn about declination and final lowering. While it is difficult to reach a definite conclusion for Standard Mandarin, the final lowering phenomenon is much more convincing in the high-level tone case in North-East Mandarin as reported by Cui and Kuang (2020). In their data, the f0 contour plummets on the final high-level tone in a statement and presents a distinct f0 lowering, in contrast to the Standard Mandarin counterpart. Interestingly, this final lowering effect only occurs on the high-level lexical tone and not the other three lexical tones.

In a more phonological description, Peng et al. (2005) posited that Mandarin varieties, including Standard Mandarin (*Putonghua*), Taiwan Mandarin (*Guoyu*), and Ronggaohua (a Jianghuai Mandarin variety), signal statements with a L boundary tone. What remains unclear is whether they intended to refer the L boundary tone to a natural global declination, a local final lowering, or a boundary tone, which, in non-tonal languages, are different concepts. In a tonal language, it is hard to tell what an intonational boundary tone is by looking directly at the f0 contour without deducting the lexical tonal contour. Therefore, it is open for further discussion whether, in tonal languages, a L boundary tone should be treated the same as in a non-tonal language; if not, how should they be analyzed differently?

In addition to f0, other cues, such as duration and intensity, are involved in signaling utterance finality of statements in Mandarin. For example, Yuan (2006) examined statements together with questions and showed that the duration of the final syllable in statements was substantially longer than the previous syllables, while the intensity showed a decreasing pattern towards the end of the statement. Whether these cues are specified in the intonation phonology, like in English and Bengali (Hayes and Lahiri 1992), or a by-product of f0 changes, calls for further research (see relevant discussion in Zhang and Lahiri, in prep).

In summary, Mandarin varieties utilize f0, duration, and intensity to mark statement boundaries. Evidence for declination is mixed. While statements have often been proposed to have a low boundary tone, whether the tone indicates utterance-level declination or final lowering localized to the boundary of an utterance remains challenging to ascertain.

### 8.3.2 Boundary of questions

In this section, we will review the local boundary phenomena for two major types of questions in Mandarin, polar question, and *wh*- question, as well as their subcategories. These are the most studied types in the prosody literature.

#### 8.3.2.1 Polar questions

Polar questions are also known as yes-no questions (henceforth *YNQ*) and can be syntactically unmarked, compared to statements. In this case, only by altering the intonation of the utterance does the speech act change from declarative to interrogative (intonational yes-no question, henceforth *IntQ*). Note that in Mandarin, polar questions can also be syntactically marked with an utterance-final question particle. One frequently used question particle is *ma* 吗,[2] often used as a neutral non-biased question marker (yes-no question with a *ma* particle, henceforth *ma-Q*). Examples of these are given below in (1-3).

[2] Polar questions in Mandarin Chinese can also end with the question particle ba 吧, which is not considered here.

(1) Statement:

| 这 | 是 | 一 | 本 | 书。 |
|---|---|---|---|---|
| *zhe51* | *shi51* | *yi51* | *ben214* | *shu55* |
| this | be | one | QUANT | book |

‘This is a book.’

(2) Intonational Question (IntQ):

| 这 | 是 | 一 | 本 | 书？ |
|---|---|---|---|---|
| *zhe51* | *shi51* | *yi51* | *ben214* | *shu55* |
| this | be | one | QUANT | book |

‘This is a book?’

(3) YNQ with a ma particle (ma-Q):

| 这 | 是 | 一 | 本 | 书 | 吗？ |
|---|---|---|---|---|---|
| *zhe51* | *shi51* | *yi51* | *ben214* | *shu55* | *ma* |
| this | be | one | QUANT | book | PARTICLE |

‘Is this a book?’

Both (2) and (3) can serve as information-seeking or rhetorical questions, though with prosodic differences. Earlier studies tend to focus more on the intonation patterns of YNQs with/without particles although these studies usually do not explicitly state the functions of the questions.

Similar to statement intonation, YNQs have also been reported to have global and local f0 effects. In contrast to the ambiguity between statements' declination and final lowering, the issue surrounding YNQs is whether there is a global or local f0 rise. The common observation is that IntQ exhibits a rising intonation pattern, while ma-Q does not always rise. Concerning IntQ, reports vary in how and to what extent the rising intonation occurs.

More specifically, Shen (1990) schematized the statement tune, IntQ tune, and ma-Q tune, which showed that both IntQ and ma-Q had higher f0 than statements at the beginning of an utterance. The ma-Q tune ended at the same low level as the statement tune while the IntQ tune ended much higher, suggesting that IntQ differs from both statement and ma-Q around the boundary. Thus, IntQ and ma-Q, despite both being YNQs, may have different intonations. Ni and Kawai (2004) replicated Shen's findings.

Liu and Xu (2005) elicited different types of YNQs using sequences of syllables with identical lexical tones under different focus conditions. Their findings suggested that the raising of f0 accelerates towards the end in questions, which led them to conclude that the f0 raising in questions is not linear, but exponential or double-exponential, and that the accelerated final rise is a core component of the global question function.

It is worth noting that IntQ in Mandarin varieties often shows a raised pitch register, though the specifics may vary due to different research scopes and methodologies For example, both Shi (1980) and Shen (1985) noted that IntQ intonation starts and ends higher than statements in Standard Mandarin and Beijing Mandarin, respectively. Yuan et al. (2002) posited a higher phrase curve for interrogatives than declaratives (without resorting to tonal specification at the boundary in modelling the two intonation types), while Zeng et al. (2004) reported a rising slope, which varies by tone.

Research also differs on where exactly the f0 raising occurs if the rise is to be taken as local. Lin (2004; 2006) showed that the registers of the starting and ending points of the last non-neutral lexical tones were higher. Lee (2005) concluded that the raising starts at the last NP of IntQs in Beijing Mandarin. Shen (1990) proposed that f0 raising starts from the first few syllables of YNQs – both IntQ and ma-Q – compared to their corresponding statement intonations.

YNQs can be both information-seeking (ISQ) and biased for certain responses from the interlocutors, such as a rhetorical question (RQ) or an echo question (see review in Chen et al. in revision.). Zahner-Ritter et al. (2022) investigated the two types with and without a final question particle and found that generally speaking, RQs showed lower f0 throughout the whole utterance. Furthermore, the f0 range of the first word was larger than the ISQ counterpart but the f0 range of the boundary was compressed.

From a phonological perspective, Peng et al. (2005) described both the ma-Q and echo IntQ (i.e., questions that echo the entire or a part of the previous statement) tunes as having a raised pitch register at the beginning of the utterance and a H boundary tone. Zhang (2018b) studied Tianjin Mandarin and also found raised pitch register for IntQ; she instead proposed a H floating boundary tone, which modifies the final lexical tone for IntQ (see Section 8.4 for a more detailed discussion on this point).

#### 8.3.2.2 Wh- questions

Wh-questions (henceforth *wh-Q*), also known as constituent questions, are questions starting with content pro-forms (more commonly known as wh-words, or interrogatives), including *what*, *when*, *where*, *who*, *whom*, *which*, *whose*, *why*, and *how*. In Mandarin, these pro-forms can appear in situ in both declarative (as indefinites) and interrogative sentences. Like YNQs, Wh-Qs can be used for seeking information or as a rhetorical question. As illustrated in (4-6), both types of wh-Q and their corresponding statements can be string identical. Listeners mainly use prosodic cues, along with contextual information, to disambiguate the sentence types.

(4) Wh-statement:

| 他 | 在 | 说 | 些 | 什么。 |
|---|---|---|---|---|
| *ta55* | *zai51* | *shuo55* | *xie55* | *shen35-me* |
| he | CONT | say | some | something |

‘He is saying something.’

(5) Information seeking wh-Q:

| 他 | 在 | 说 | 些 | 什么? |
|---|---|---|---|---|
| *ta55* | *zai51* | *shuo55* | *xie55* | *shen35-me* |
| he | CONT | say | some | what |

‘What is he saying?’

(6) Rhetorical wh-Q:

| 他 | 在 | 说 | 些 | 什么? |
|---|---|---|---|---|
| *ta55* | *zai51* | *shuo55* | *xie55* | *shen35-me* |
| he | CONT | say | some | what/something |

‘WHAT (exactly) is he saying?’

There are few studies on the prosody of wh-questions in Mandarin. Yang et al. (2020) compared wh-declarative and information-seeking wh-Q and reported a significantly higher f0 of the wh-word in wh-Qs (compared to wh-statements). Interestingly, they also found that the f0 remained higher until the end of the utterance. The f0 range also differed in the wh-word and continued until the end of utterances. Zahner-Ritter et al. (2022) compared information seeking wh-Q and rhetorical wh-Q and found that rhetorical wh-Qs were generally higher in f0, except for parts of the verb.

From the phonological perspective, as these findings indicate an overall register difference rather than any specific salient rise localized to the boundary position, researchers typically avoid positing a boundary tone but the question remains whether the f0 raising, like in YNQs, may be taken as the phonetic realization of a boundary H tone.

Other than f0, durational differences were also reported. According to Yang et al. (2020), the average duration of wh- statements is longer than wh-Qs. Zahner-Ritter et al. (2022) found that rhetorical questions were longer than information-seeking ones. They also reported voice quality differences, with rhetorical questions more often showing non-modal voice quality (glottalized voice) than information-seeking questions.

Before we end this section, one thing of particular note is the need for a more refined categorization of YNQs and wh-Qs in terms of their illocution types, as in Zahner-Ritter et al. (2022). More research is needed to understand the nuances that intonation, together with various particles, can convey in different communicative contexts and reflect the prior beliefs, contextual evidence, and attitudes of interlocutors.

### 8.4 Types of boundary phenomena

In this section, we start with the assumption that the final f0 raising/lowering or rising/falling is the phonetic manifestation of a phonological boundary tone in Mandarin intonation. We will show, in Fig. 8.1, that in Mandarin, localized f0 modifications of lexical tones at intonation boundaries may be categorized into two types: One type involves f0 modifications that respect the pitch contours of lexical tones (hereafter referred to as modifying boundary tone); and the other type involves significant lengthening of the segmental syllable as an additional docking site for intonational pitch events, which we will refer to as additional boundary tone hereafter.

It is important to note that even in the case of an additional boundary tone, the original lexical tone's f0 shape remains largely intact and exerts a significant influence over the f0 realization of the additional boundary tone. If pressed to draw an analogy between the two types of boundary tones and segmental variations, we may compare modifying boundary tone to nasalized vowels and additional boundary tone to a vowel + nasal sequence.

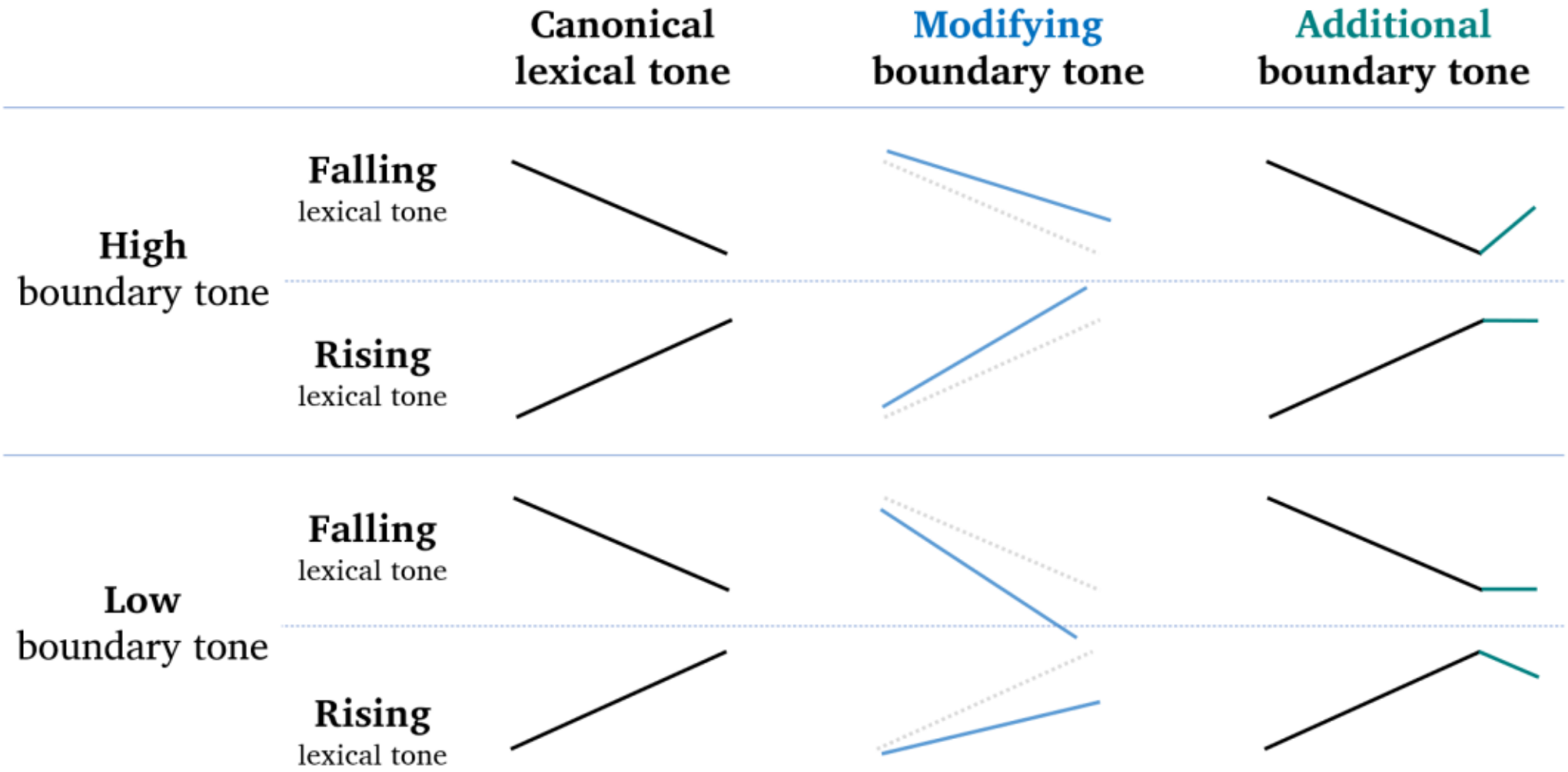

**Figure 8.1:** Schematic representation of canonical lexical tones and how two different types of boundary tones shape the lexical tones. The dotted lines represent in canonical lexical tones.

<insert Figure 8.1 here>

The modifying type of boundary tone has also been analyzed phonologically as a "floating" boundary tone in Zhang (2018b), where a H floating boundary tone induces a slower falling slope in a falling lexical tone and a sharper rising slope in a rising lexical tone without having to introduce a target as in the case of an additional boundary tone.

In the literature, Chao (1933) has described boundary-related tonal phenomena as "simultaneous addition" (the modifying type) and "successive addition" (the additional type). Drawing on the possibilities discussed in Hyman and Monaka (2008), the modifying type may be referred to as "submission" ("the intonational tones invade and override the lexical tones"), where the lexical tone gives way to the intonation tone, and the addition type as "accommodation" ("the terrain is divided up somehow such that the lexical and intonational tones minimally interact"), with tones of different levels co-existing. However, the interpretation of "submission" may vary among researchers: it could refer to situations where intonation modifies the lexical tone slightly without neutralizing any tonal contrast, as in Zhang's (2018b) analysis; or, it could refer to cases where tonal categories are merged due to intonation influence. For example, in Cantonese, the H boundary tone in questions almost neutralizes the tonal contrasts between the low tones (21, 22, 23), and the high-rising tone (25) (Ma et al. 2006), making it difficult for listeners to accurately differentiate the lexical tones in this condition (see also discussion in Chen 2022b).

### 8.4.1 Modifying boundary tone

The existing literature suggests that the primary cues for boundary intonation in Mandarin may be classified as the modifying type. In addition to Standard Mandarin (e.g., Lin 2004), Tianjin Mandarin is another dialect reported to have a similar phenomenon (Zhang 2018b), as shown by the stylized lines in Figure 8.2. None of the lexical tones undergoes overall contour changes. A falling tone remains falling, and a rising tone remains rising, but the degrees of rises and falls change to signal the statement-question intonation contrast. To our knowledge, modifying boundary tones has only been reported in questions in Mandarin varieties. Such a strategy is likely to be used in other types of intonation in natural speech. However, since the perceptual difference introduced by a modifying boundary tone can be relatively subtle, it may be easily masked by the lexical tone contours.

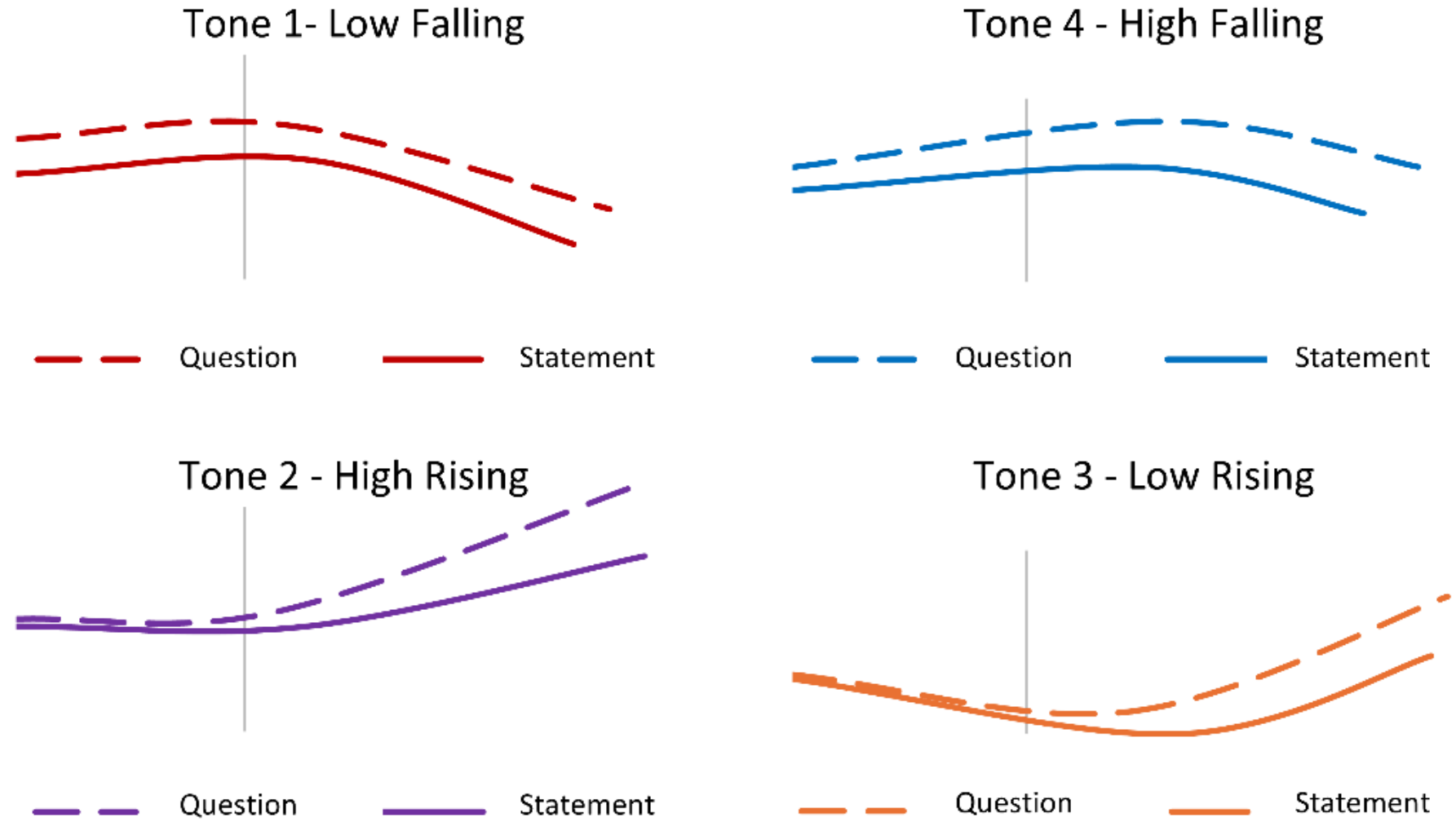

**Figure 8.2:** Stylized lines for statement tune (Statement, solid lines) and intonational question tune (Question, broken lines) in Tianjin Mandarin (adapted from Zhang 2018b: 83). The vertical lines indicate boundaries between onsets and rhymes.

<insert Figure 8.2 here>

### 8.4.2 Additional boundary tone

Additional boundary tones are rarely reported for intonation in Mandarin Chinese, partly because most research has focused on the linguistic functions of intonation, such as marking statements vs. questions. However, when one ventures into intonation used for socio-cognitive functions, such as expressing emotions or signaling attitudes, additional boundary tones abound. Chen (2022a) discussed such a case in Standard Mandarin, with the H lexical tone as an example which can have an additive falling contour at the end when spoken with anger/annoyance. Muller-Liu (2006; 2018) discussed Chao's "successive addition" concept and showed examples of falling edge tones in Mandarin. They also collected attitudinal ratings of the example sentences and showed that a falling edge tone can be a sign of either positive or negative emotion.

In the following section, we present a more extensive range of corpus data to illustrate the presence of additional boundary tones in Mandarin Chinese, aiming for a more systematic documentation of the phenomena. The data comes from recordings of TV shows, including reality programs and comedy shows, sourced from online media. These recordings were made using a mobile phone, capturing internal device sounds without external interference. Since many studies (e.g., Fuchs and Maxwell 2016; Zhang et al. 2021; 2024) have shown f0 to be a robust acoustic measurement that can be reliably tracked by various recording devices and formats as long as the compression is not extreme, we believe the f0 contours in these recordings are largely reliable.

Here we will focus on cases where additional boundary tones are used, often to convey correction, emphasis, objection, impatience, or *sajiao* (a form of being cute while requesting something). However, we will not go into detail about their specific linguistic and paralinguistic functions due to both space limit and our lack of understanding of the full range of meanings these additional tones express

Form-wise, the additional boundary tones can be as simple as a rise or dip in f0, or they may involve more complex patterns such as rising and falling multiple times. For descriptive clarity, we define the number of tones at the boundary position by assuming that each f0 peak/valley represents a pitch target as the surface realization of one phonological tone, in line with the assumptions of the Autosegmental-Metrical tradition (Ladd 1996; Pierrehumbert 1980). Thus, a simple rise or fall is a mono-tonal H or L boundary tone; a rise-fall is a bi-tonal HL boundary tone; and anything more complex will be a multi-tonal boundary tone.

### 8.4.2.1 Mono-tonal additional boundary tones

In this section, we will introduce how additional boundary tones can be realized after different lexical tones. Examples hereafter are presented in Chinese text first and then pinyin transcription with tones represented in Chao letters. The target syllables for illustration are marked in bold.

In Standard Mandarin, an additional L boundary tone is observed in a chanted call tune, as shown in (7). The speaker asks the listener to be louder and, therefore, produces a chanted call tune with an imperative utterance. The final syllables carry a high-level tone, Tone 1 [55].

(7) **Tone 1 [55]** + *L boundary tone*

| 大 | 点 | 声 |
|---|---|---|
| *da51* | *dian214* | ***sheng [55]*** *+ L* |
| big | a-little | voice |

'Be louder'

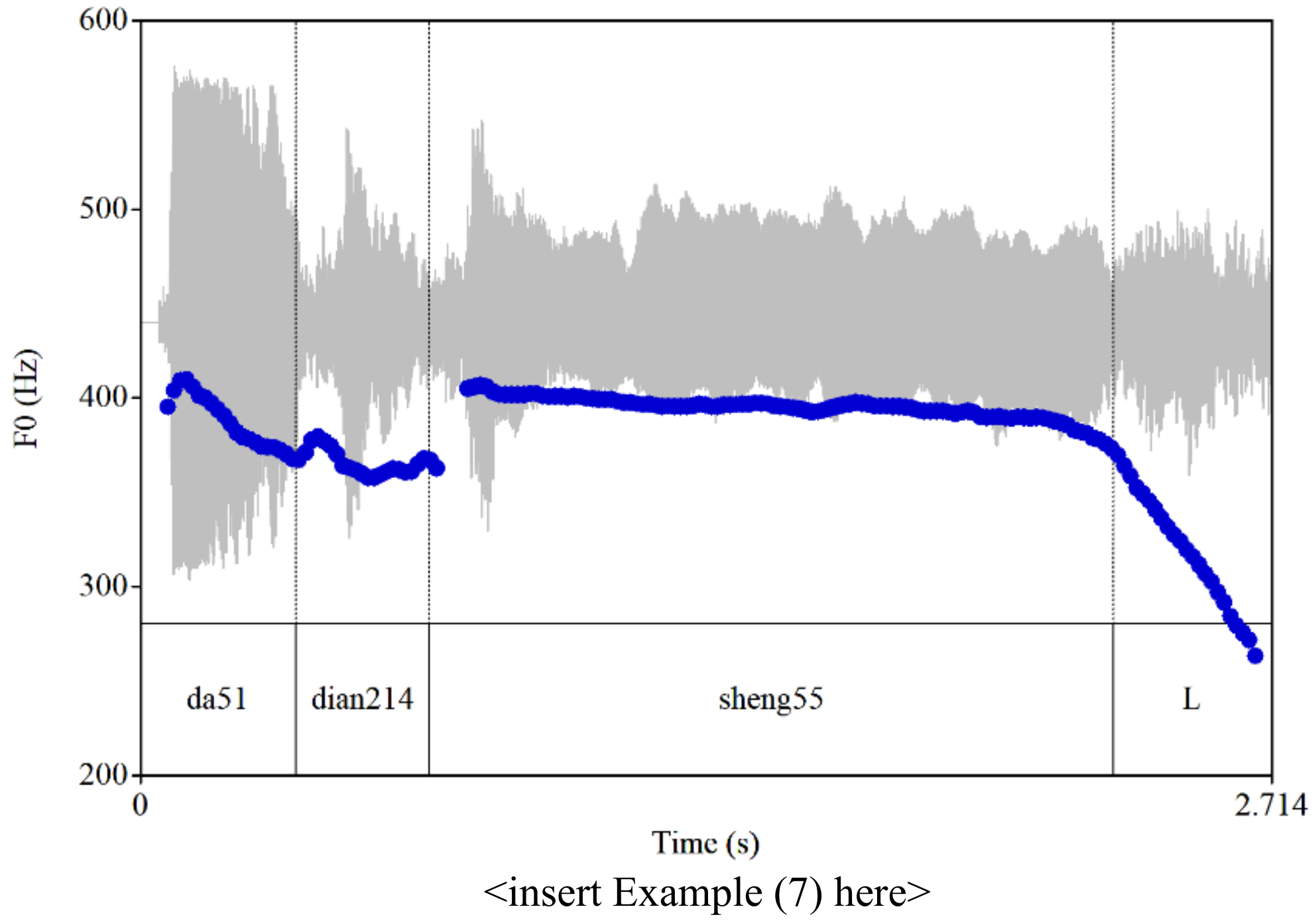

<insert Example (7) here>

(8) is a response to (7), which asks the interlocutor to repeat more loudly. This example carries a falling tone (Tone 4 [51]) on the final syllable. Identifying a L boundary tone following a falling tone, as this example shows, is not straightforward. We may take the final f0 drop as the realization of this L boundary tone. Admittedly, the f0 manifestation of the L tone here is rather subtle and brief, potentially raising questions about the L tone identity.

(8) **Tone 4 [51]** + *L boundary tone*

| 您 | 上 | 哪 | 去 |
|---|---|---|---|
| *nin35* | *shang54* | *na214* | ***qu[51]*** + *L* |
| you | towards | where | go |

'Where are you going'

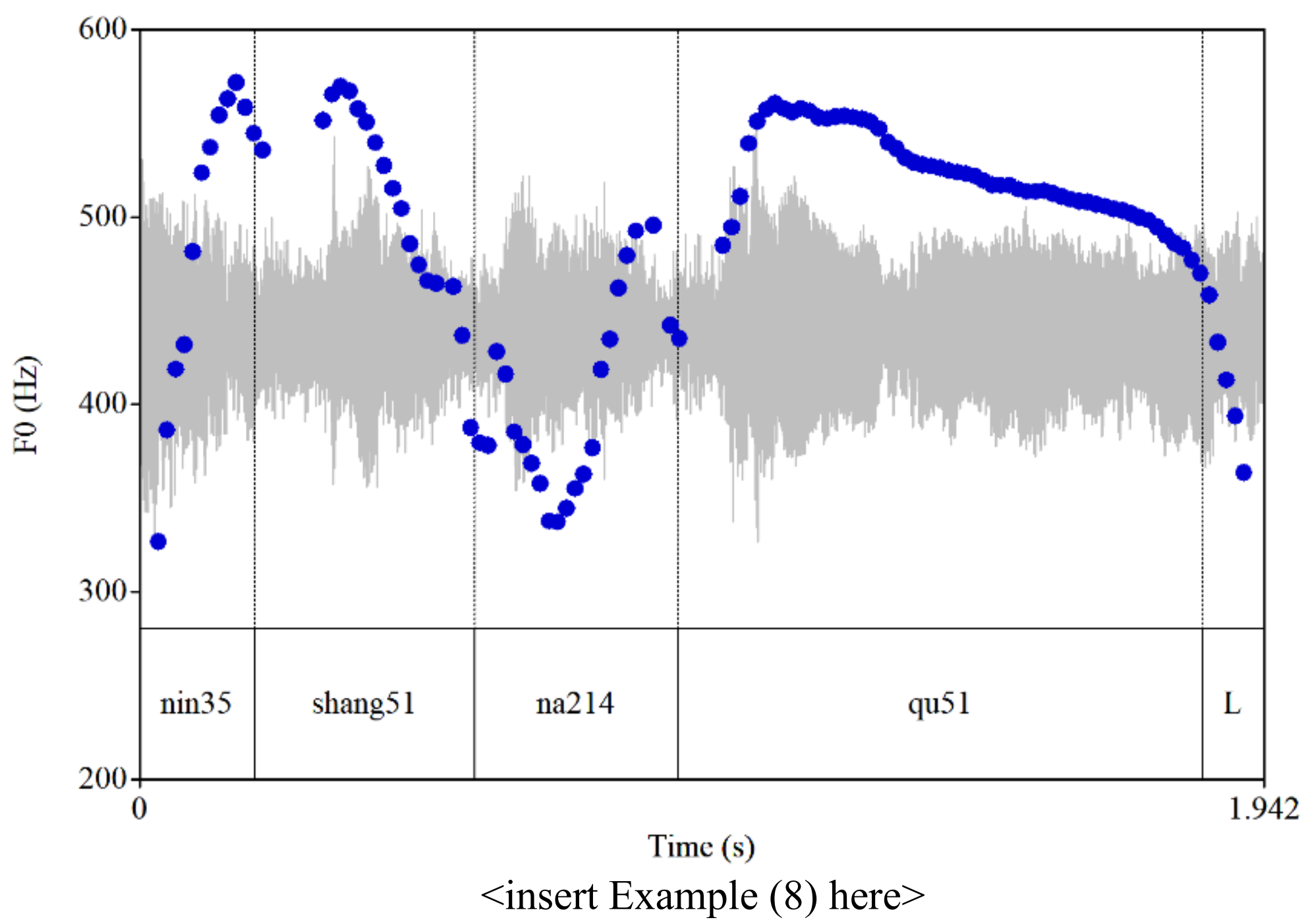

<insert Example (8) here>

In (9), a comedian from a *xiangsheng*[3] comedy duo starts the show by introducing himself and then his partner in the following utterance. Both utterances end with a rising tone, Tone 2 ([35]). Both utterances show a clear drop at the end of the rising lexical tone, suggesting a L boundary tone.

(9) **Tone 2 [35]** + *L boundary tone*

| | | | | | |
|---|---|---|---|---|---|
| （我叫）孟 | 鹤 | 堂。 | （…我的搭档叫）周 | 九 | 良。 |
| *Meng51* | *He51-**Tang[35]*** + *L* | | *Zhou55* | *Jiu214-**Liang[35]*** + *L* | |
| SURNAME | GIVEN-NAME | | SURNAME | GIVEN-NAME | |

'(I am called) Meng He-Tang. (This is my partner, called,) Zhou Jiu-Liang'

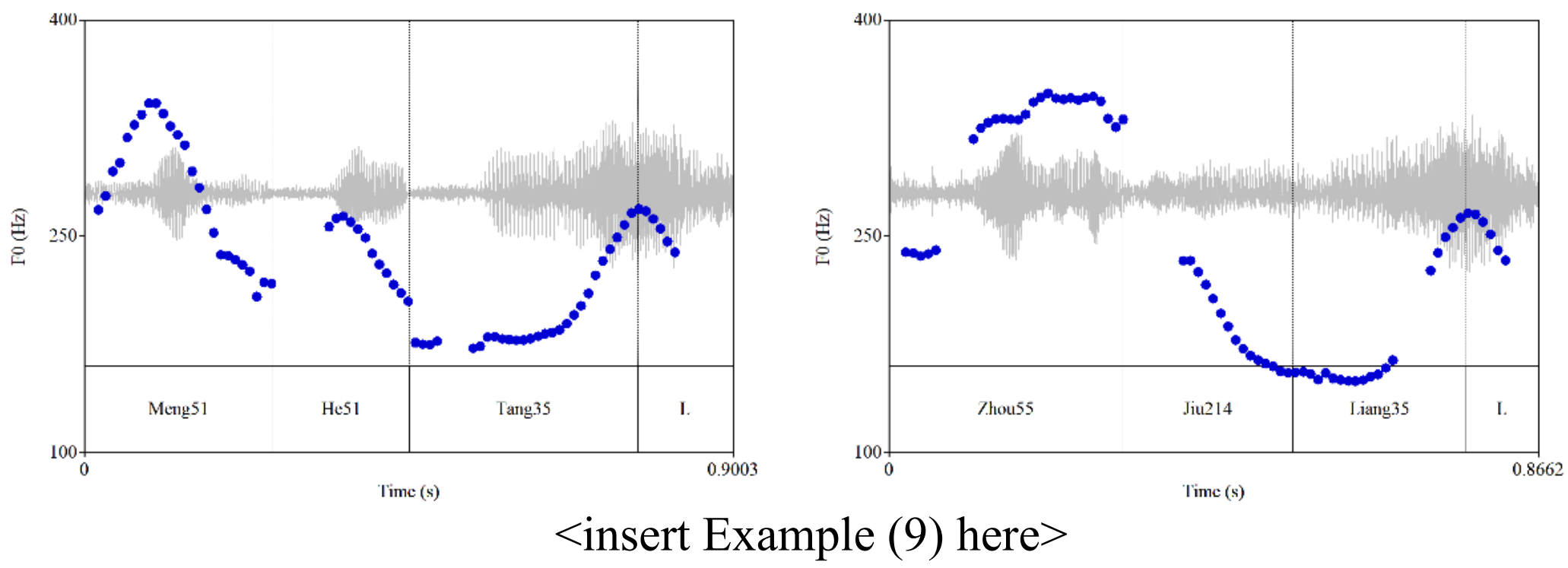

<insert Example (9) here>

[3] A Chinese stand-up comedy form, typically involving two comedians bantering with each other.

(10) is an example of how a L boundary tone manifests on a fall-rise tone, Tone 3 [214]. The Tone-3 syllable does not reach the target H tone in the canonical lexical tone, and the f0 drops at the end of the utterance. In this context, the speaker (A) was asking their interlocutor (B) to hand over what B took from A.

(10) **Tone 3 [214]** + *L boundary tone*

你 给 我

*ni214 gei214* ***wo[214]*** *+ L*

*you give me*

'Give it to me'

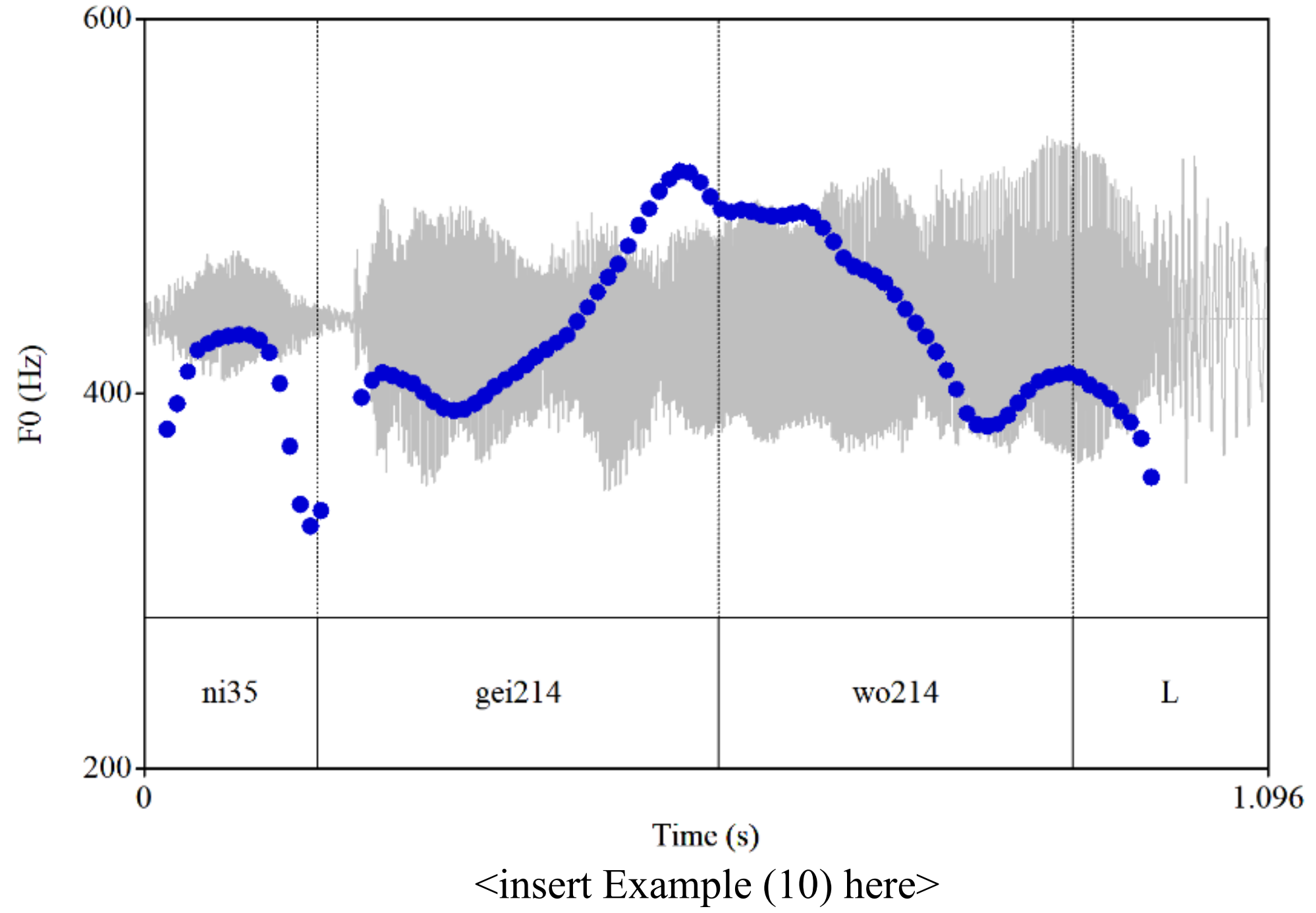

<insert Example (10) here>

### 8.4.2.2 Bi-tonal additional boundary tones

In non-tonal languages, it is not uncommon to see more than one tone at the boundary. For instance, in Bengali, it is possible to have both a phonological phrase tone and an intonational phrase tone consecutively, or a bi-tonal intonational phrase boundary tone (Hayes and Lahiri 1991). Similarly in Korean, a boundary tone cluster can include as many as five tones, such as an LHLHL boundary tone (Jun 2000). However, it is rare to see such additional tonal clusters in Mandarin varieties, since keeping the identity of lexical tones seems to be a top priority. In this section, we present cases of a bi-tonal HL boundary tone annexed to the four lexical tones respectively.

In (11), the speaker tried to call an aunty from afar. The lexical high-level Tone 1 [55] is realized with a bi-tonal HL boundary tone, which raises the f0 of the level tone first, and then drops at the end.

(11) **Tone 1 [55]** + *HL boundary tone*
姑！
***gu [55]*** *+ HL*
aunt

'Aunty!'

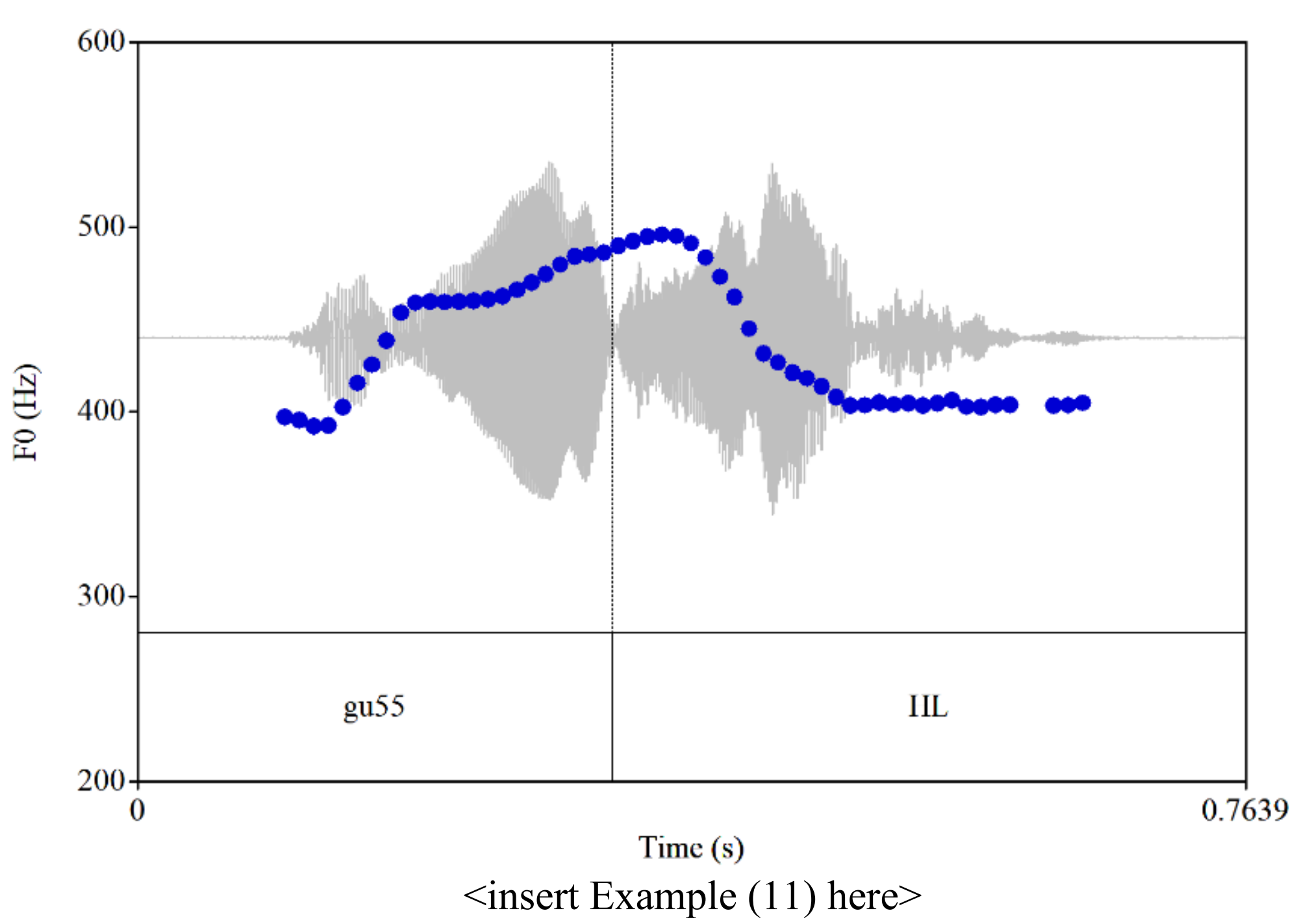

<insert Example (11) here>

The HL boundary tone is easier to identify with a falling tone, Tone 4 [51], as illustrated in (12), where the lexical falling tone ends in a low f0 target but then rises and falls again at the end. In this utterance, the speaker corrects a friend's choice of a Chinese character. Such additional tonal targets are different from lab speech reported in Chen and Gussenhoven (2008) where distinctive realizations of lexical tonal contours have been found.

(12) **Tone 4 [51]** + *HL boundary tone*

| (是) | "孟 | 鹤 堂" | 的 | "孟"。 |
|---|---|---|---|---|
| | *Meng51* | *He51-Tang35* | *de* | ***Meng[51]***+ *HL* |
| | surname | given-name | POSSESSIVE | surname |

'It is the "Meng" in "Meng Hetang".'

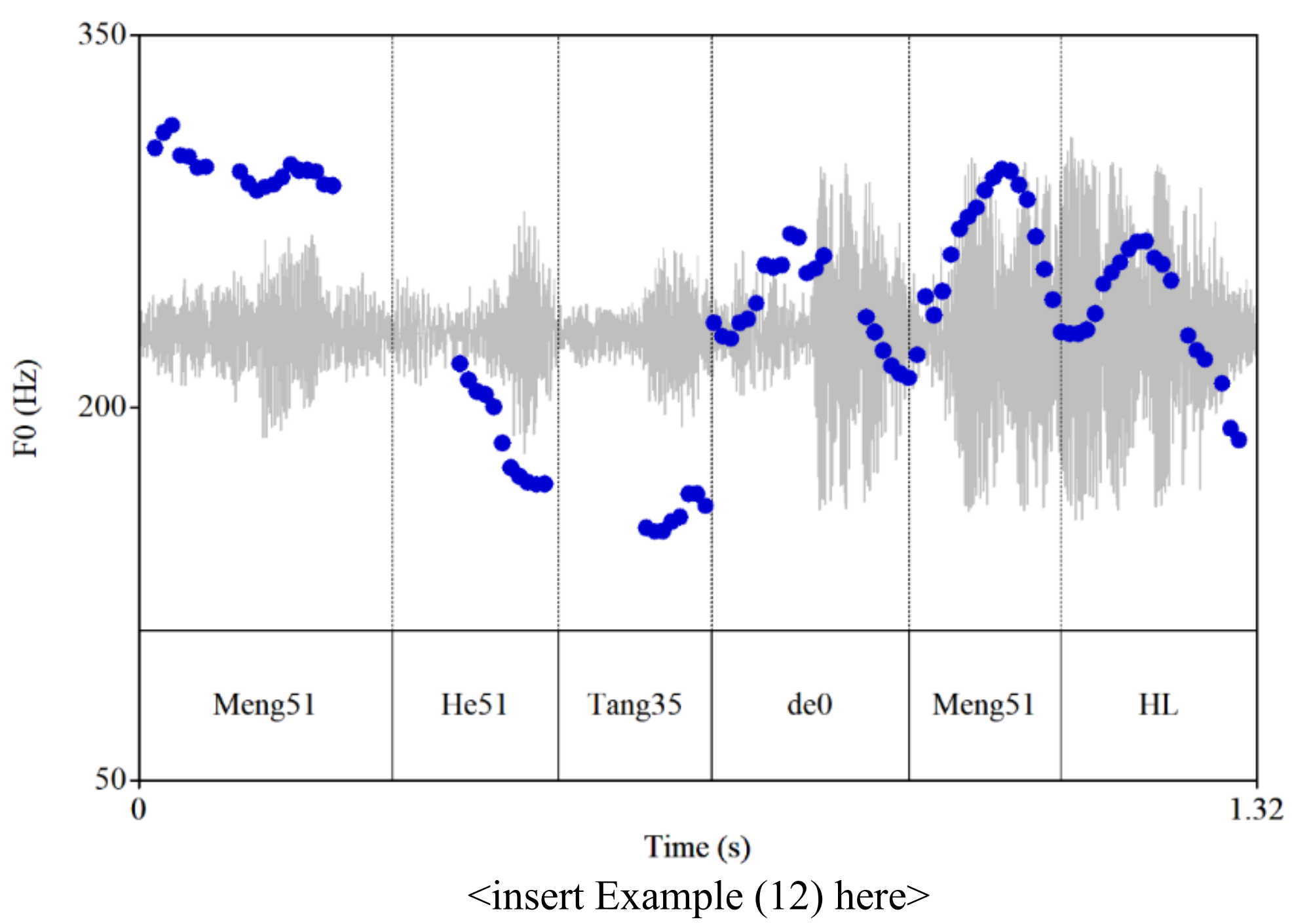

<insert Example (12) here>

The HL boundary tone is more difficult to identify when added to the lexical Tone 2 [35] or Tone 3 [214] since both end high. In (13), an HL boundary tone is posited given the f0 rises after the T3 over "*gou* [214] and before the falling contour, in contrast to the much lower f0 following "*wo* [214]" in (10).

(13) **Tone 3 [214]** + *HL boundary tone*

| 再 | 来 | 是 | 小 | 狗 |
|---|---|---|---|---|
| *zai51* | *lai35* | *shi51* | *xiao214* | ***gou[214]*** + *HL* |
| again | come | be | little | dog |

'If I visited again, I'd be a doggie.'

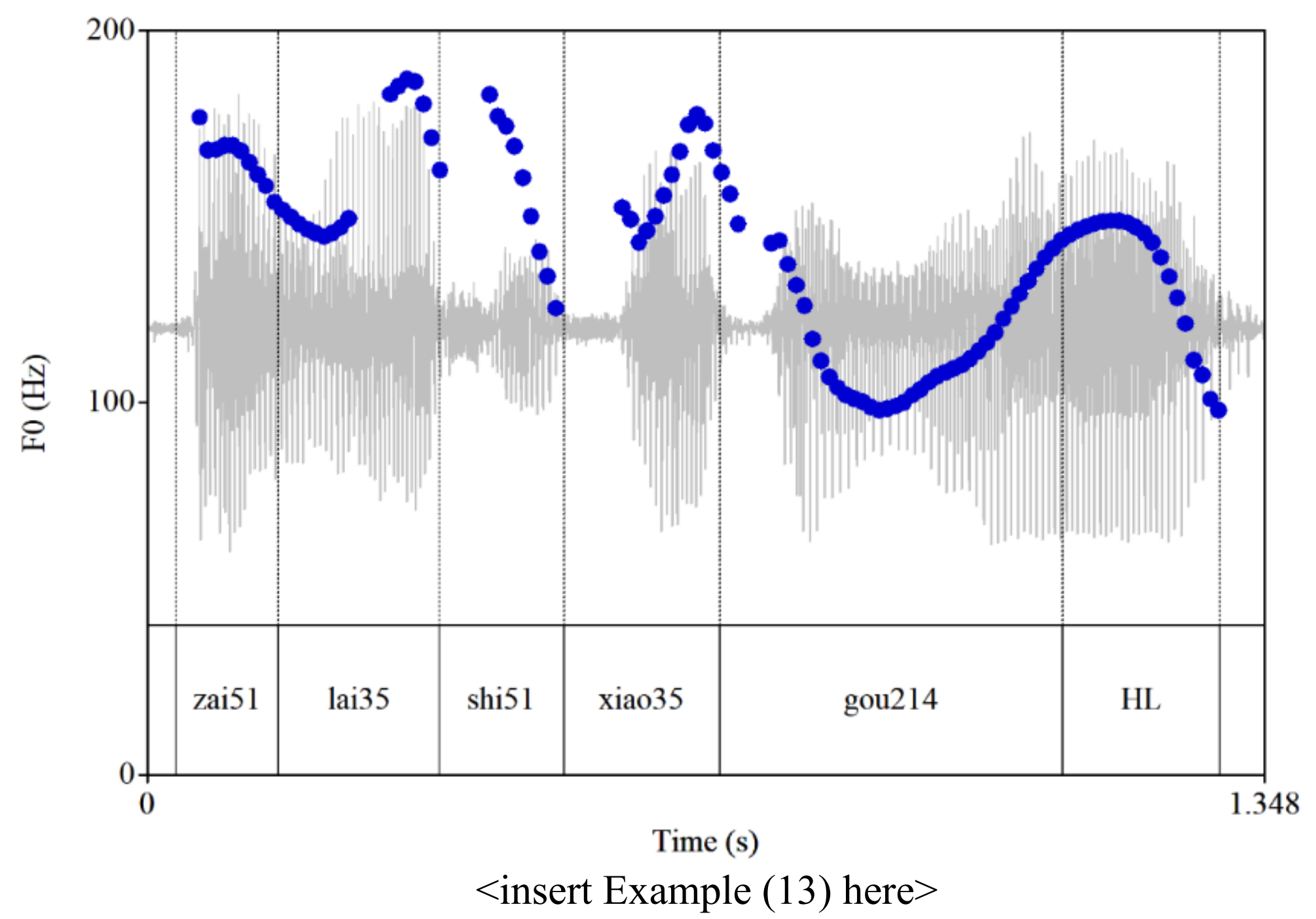

<insert Example (13) here>

Similarly, in (14), an HL boundary tone is posited given that the f0 after the rising Tone 2 *yu* [35] is much higher before the fall, compared to that in (9), which ends low without a salient f0 peak. In (14), the speaker makes fun of how people, especially poets, often wish to be something else, while being human is actually a significant privilege.

(14) **Tone 2 [35]** + *HL boundary tone*

| （我要当） | 一 | 条 | 鱼 |
|---|---|---|---|
| | *yi51* | *tiao35* | ***yu[35]*** + *HL* |
| | one | QUANT | fish |

‘(I want to be) a fish’

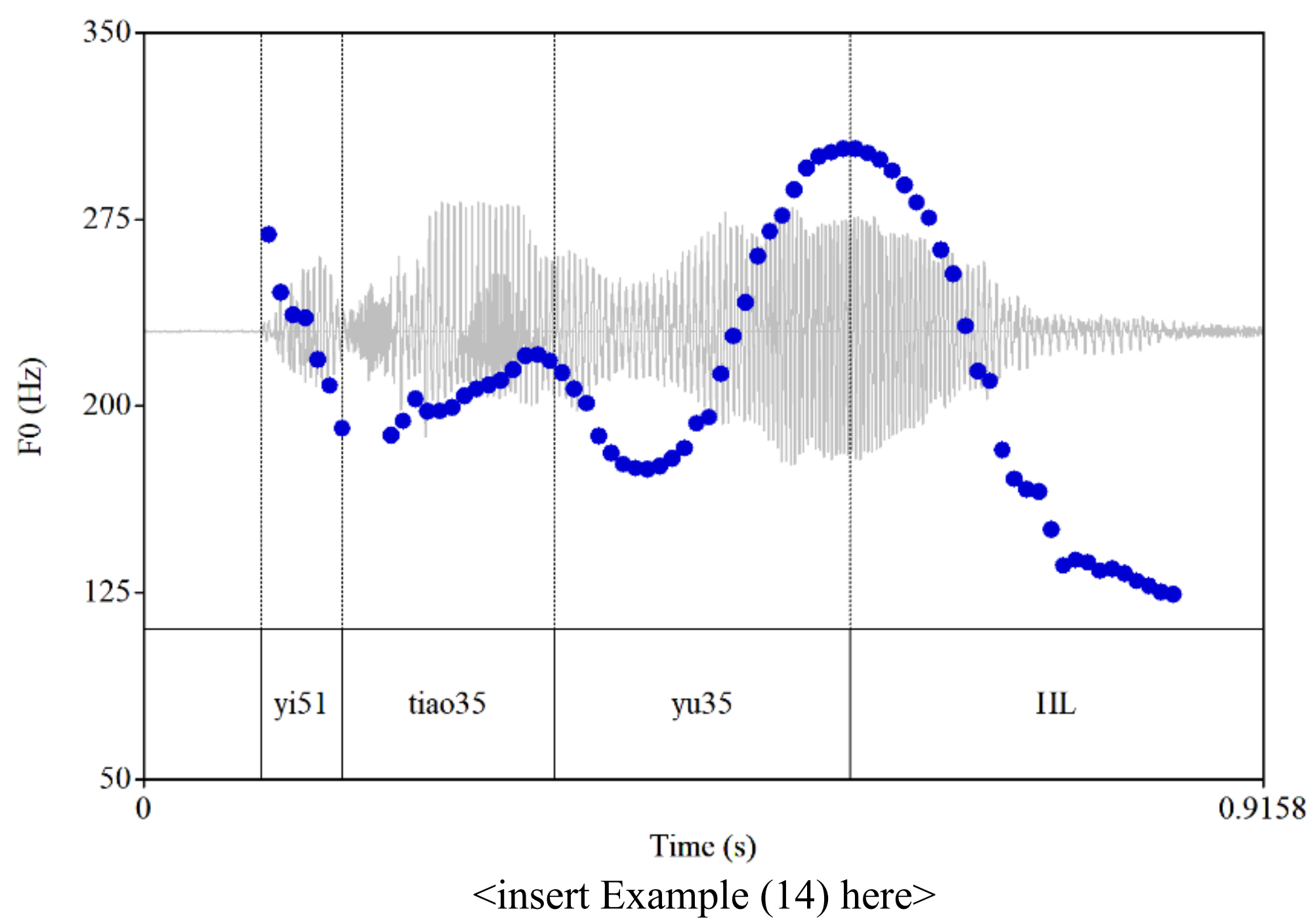

<insert Example (14) here>

For a neutral-tone syllable in (15), the presence of an HL boundary tone causes the intonational tone to stay nearly as high as the previous HL tone and then end with a low-pitch target. This necessitates an HL boundary tone rather than an L boundary tone, as the f0 rises slightly on the second neutral-tone syllable before falling. In this scenario, the speaker is begging his teammates to agree with him.

(15) **Neutral tone** + *HL boundary tone*

| 同意 | 了 | 吧 |
|---|---|---|
| *tong35-yi51* | *le* | ***ba*** + *HL* |
| agree | PART | PART |

'Just agree with it.'

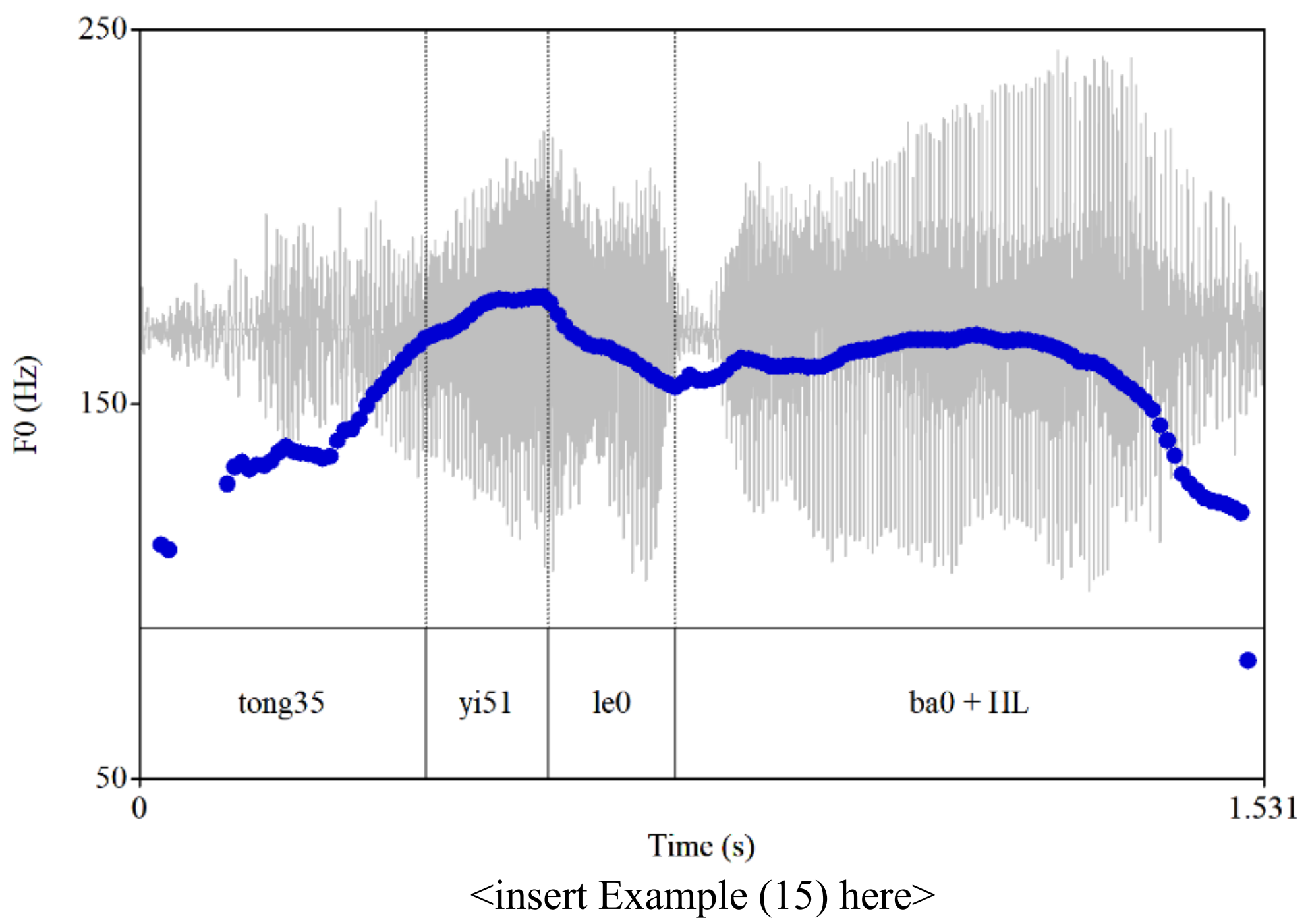

<insert Example (15) here>

#### 8.4.2.3 Multi-tonal additional boundary tones

Bi-tonal boundary tones, though rare, are not the limit in Mandarin Chinese. In this section, we present examples of a multi-tonal boundary tone HLHL.

In (16), which follows the utterance in (12) (provided in bracket here), the speaker tries to correct the interlocutor's choice of the Chinese character in the first clause and provides the correct character in the second clause in brackets. In the first clause, the speaker uses a quadri-tonal HLHL boundary tone (i.e., two falling contours) after a lexical falling tone.

(16) **Tone 4 [51]** + *HLHL boundary tone*

不 是 这 个 "梦"。 （是"孟鹤堂"的"孟"。）

*bu35 shi51 zhe51 ge* ***Meng[51]*** + *HLHL*

not be this QUANT SURNAME

'Not this "Meng". (It is the "Meng" in "Meng Hetang").

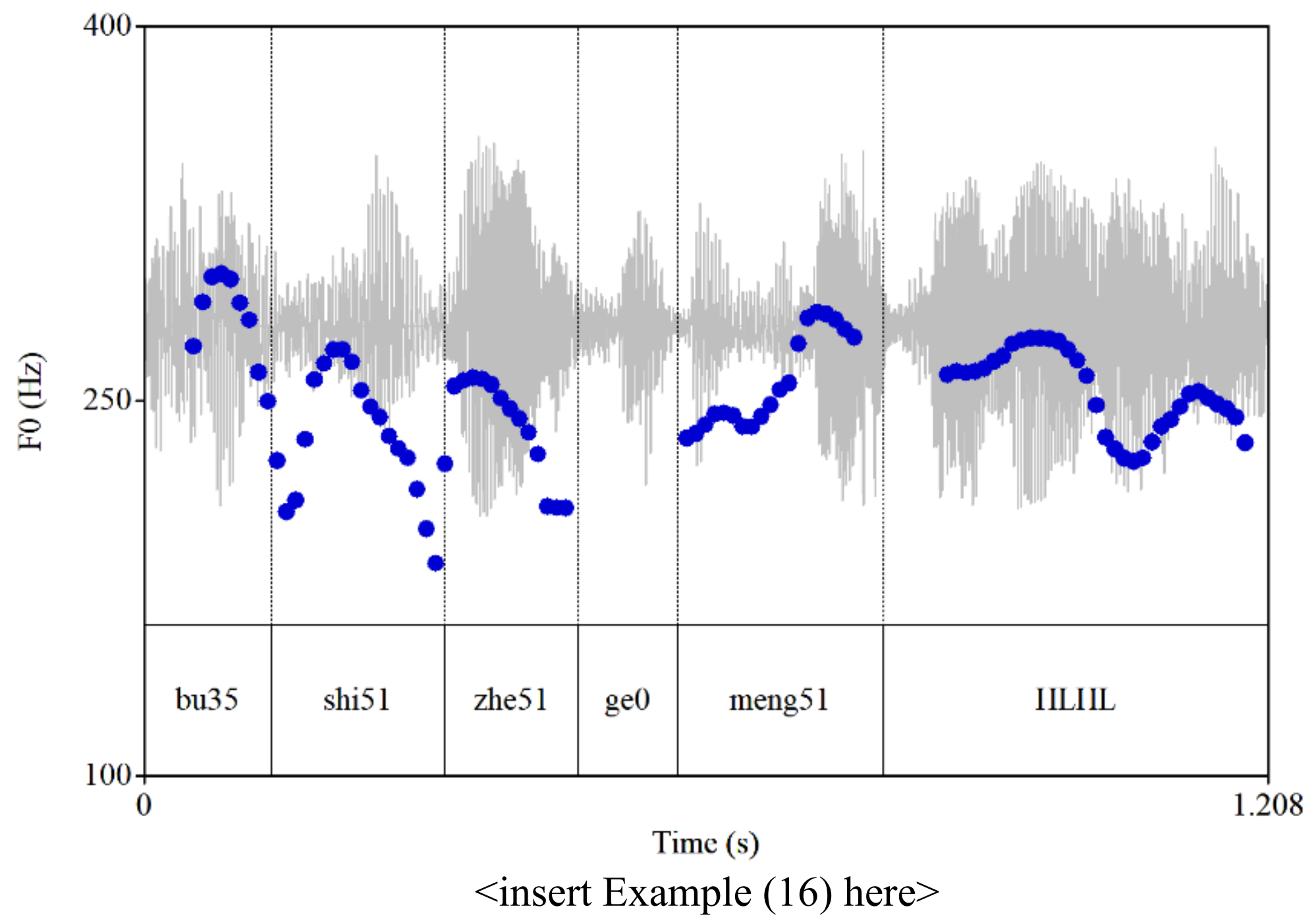

<insert Example (16) here>

In (17), we observe the HLHL boundary tone after a rising lexical tone. The context is that a comedy duo is discussing whether they can address each other using one or three characters. The speaker proposes using two characters, but his partner disagrees, thinking it is not possible. The speaker insists that using two characters is fine. The final lexical rising tone, *xing* [35], rises higher than all other tonal targets over the utterance. This is followed by a significant fall, another rise, and a final fall.

(17) **Tone 2 [35]** + *HLHL boundary tone*

| 俩 | 字 | 行 |
|---|---|---|
| *lia214* | *zi51* | ***xing[35]*** + *HLHL* |
| two | character | ok |

'(Using) two characters is OK.'

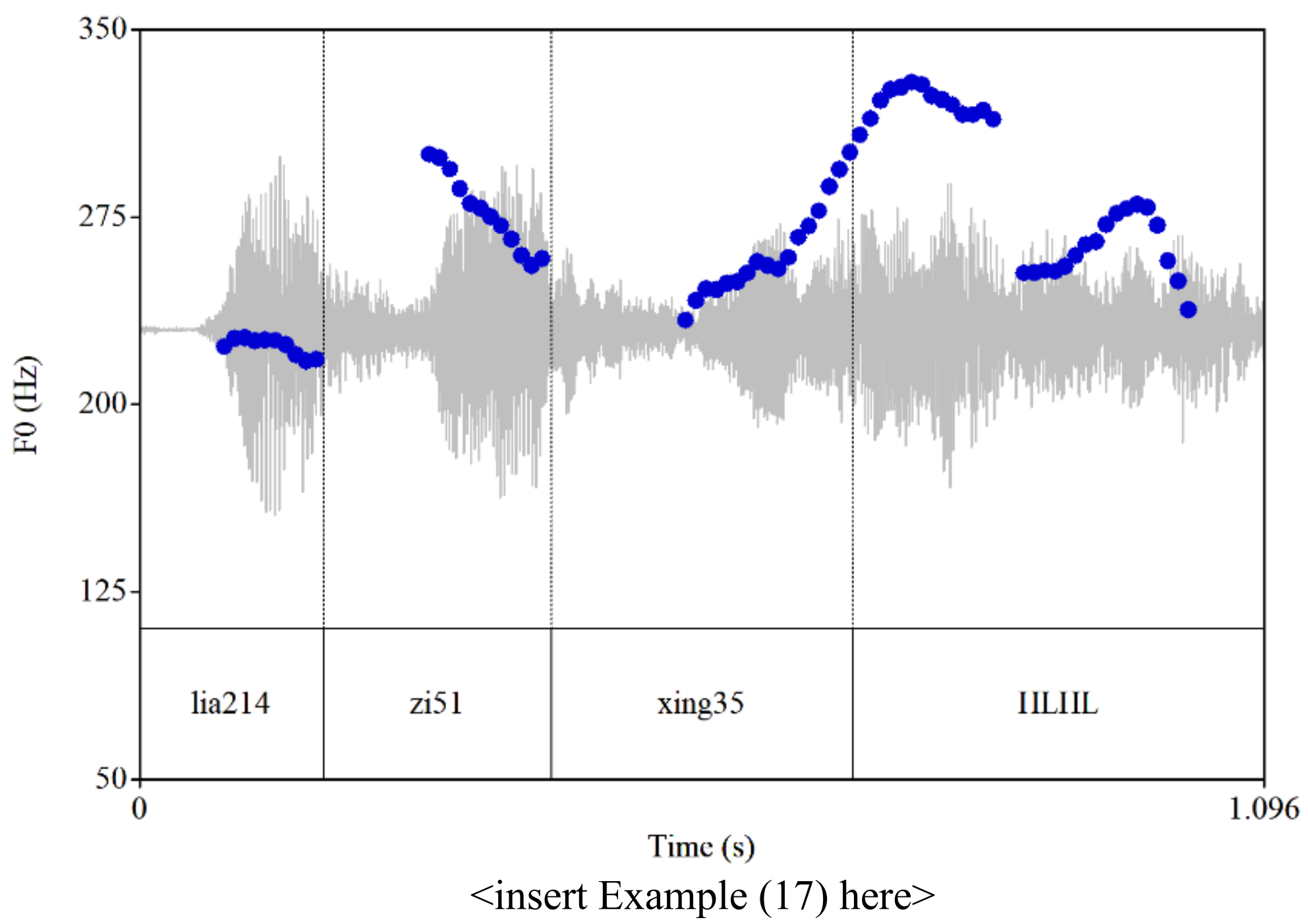

<insert Example (17) here>

To summarize the three types of additional boundary tones, we can schematize their manifestations on different lexical tones as shown in Figure 8.3.

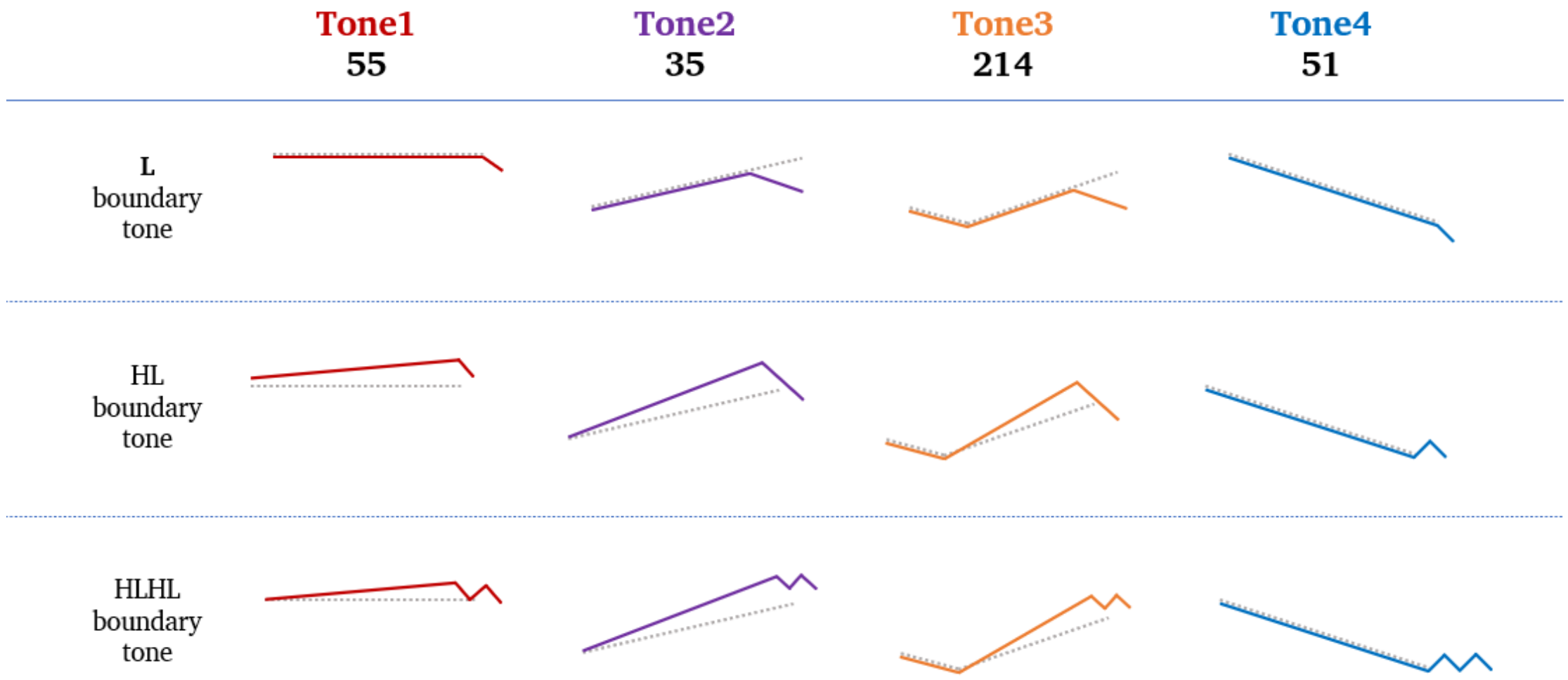

**Figure 8.3:** Canonical lexical tones and how two different types of additional boundary tones shape the lexical tones. The dotted lines are the canonical lexical tones.

<insert Figure 8.3 here>

While f0 is where the different boundary tones manifest, accompanying durational differences are also worth noting. Zhang (2018b) observed a longer duration in both IntQ and calling tunes compared to their statement counterparts. However, the sources of these durational differences seem to differ. The longer duration in the IntQ tune was attributed to the inclusion of a floating boundary tone. While this tone lacks a separate tonal target, its realization brings about durational differences. Removing the durational differences in the IntQ tune would not result in a change in its tune perception. In contrast, Zhang (2018b) reported a significant lengthening effect in the calling tune relative to the statement tune. In this case, the absence of the durational cue would result in a perceptual differences in the tune.

### 8.5 Models and approaches of intonation research

Existing studies often use different frameworks and analysis methods to investigate the same question, resulting in discrepancies. For instance, some studies on the boundary tones of questions adopt the theoretical models such as the Autosegmental-Metrical framework (Ladd 1996; Pierrehumbert 1980) to analyze discrete boundary tones, while others statistically model the continuous f0 signal as curves (e.g., using methods like GAMM and fPCA, see Section 8.5.2). This section reviews relevant theoretical and statistical models to help establish common ground for further comparisons and discussion.

#### 8.5.1 Theoretical models

Prosody researchers have proposed various models with different approaches to studying intonation. Broadly speaking, they can be classified into “concrete” and “abstract” models, as noted early on by Ladd and Cutler (1983). The “concrete” models focus on phonetic details to establish a direct mapping between the physical forms (such as pitch, intensity, and duration) and the functions of prosody. The “abstract” models, on the contrary, hold the view that abstract phonological categories and structures of intonation mediate the link between phonetic substance and meaning. Many of the existing models have been reviewed by Arvaniti (2011) and Xu (2015) (see also Zhang 2018b: 8 for a summary.) For more recent reviews of these models by their originators/developers in light of current prosody research, see Barnes and Shattuck-Hufnagel (2022).

One of the central debates, first brought up by Bolinger (1951), revolves around whether prosodic models should be configuration-based or level-based. Configurational models, such as PENTA – Parallel Encoding and Pitch Target Approximation (Xu 2005), suggest a direct link between form and function; however, they cannot easily account for utterances with the same intonation but different surface forms due to various factors such as different lexical stresses or constituent lengths. In contrast, level-based models, such as Autosegmental-Metrical Model (Ladd 1996; Pierrehumbert 1980), use a sequence of abstract high vs. low level tones to represent the whole intonational tune.

A second issue concerns the linearity or superposition nature of intonation contours. Since level-based models are largely linear, this division is more relevant for configurational models. Models such as INTSINT – International Transcription System for Intonation[4] (Hirst and Di Cristo 1998) and OXIGEN – Oxford Intonation Generator (Grabe et al. 2003) assume a linear concatenation of configurational intonation contours. In contrast, superposition models assume that surface f0 contours consist of different layers of tonal specification. Chao (1930; 1968) first compared intonation to "small ripples riding on big waves". The "small ripples" refer to lexical tones, and "big waves" to intonation. Gårding's (1983; 1987) model specifies local pitch minima and maxima, with "baseline" and "topline" connecting these points to form statement lines (Bruce and Gårding 1978). Similarly, "focal lines" connect word accent maxima and minima for focused words, forming a "grid" that describes the overall intonation contour shape. The Fujisaki model (1983) decomposes intonation into local accent commands and phrase commands, using over 10 cues including f0, amplitude, and timing, to synthesize prosodic contours. The more recent one, PENTA (Xu 2005), combines functions with different levels of prosodic information, and may be considered "quasi-superpositional" since "different communicative functions are encoded by modifying local pitch target parameters" (Xu 2015: 182). Unlike PENTA, which uses a uni-directional approach and only accounts for carry-over tonal coarticulation, Stem-ML (Kochanski and Shih 2003) is a superpositional model which assumes bidirectional smoothing, accommodating both carry-over and anticipatory tonal coarticulation.

Many of the above-mentioned models are developed based on non-tonal languages. Adding an extra layer of lexical tones in the acoustic signal certainly presents additional difficulties in applying these models. Nevertheless, attempts have been made to utilize these models for Mandarin Chinese.

Most configuration models have been used for "analysis by synthesis" (see Hirst 2011: 58) for Mandarin intonation. The models are used for generating predictive values from the raw acoustic data and using the predicted values in synthesis for evaluation.

[4] Strictly speaking, INTSINT is a transcription system rather than a theoretical model. However, it has its own theoretical assumptions, so it is included here as a theoretical model.

Since PENTA is developed based on Mandarin Chinese, it is no surprise that PENTA works well for analyzing and synthesizing tones and intonation in Standard Mandarin (Xu 2004) and other varieties of Mandarin.

Momel (Hirst and Espesser 1993; Hirst 2005) makes use of f0 extrema and models the intonation contour in a stylized fashion. INTSINT, as a transcription system for Momel, uses a set of 8-letter labels to transcribe the f0 contour directly. The labels include three absolute tone labels, t(op), m(id), and b(ottom), and five relative tone labels, h(igher), s(ame), l(ower), u(pstepped) and d(ownstepped). The absolute tone labels are defined based on a speaker's pitch range, and the relative labels are defined in relation to the immediately preceding tonal target. Zhi et al. (2016) transcribed and synthesized Standard Mandarin successfully. However, the amount of information given in this paper was limited, and it is difficult to say whether reliable labels can be given in a tonal language under the framework of INTSINT.

Fujisaki et al. (2005) also analyzed and synthesized the f0 contour in Standard Mandarin by adapting the existing Fujisaki model with an added extension specifying the lexical tones. This method has also been tested in Cantonese in Gu et al. (2007).

Different from configurational models, the linear level-based Autosegmental-Metrical model (AM) (Pierrehumbert 1980), focuses more on the phonological analysis of intonational tunes. AM is currently the most widely adopted phonological model for intonation research. It comes with an annotation system, Tone and Break Indices (ToBI), which annotates phonologically meaningful events such as pitch accents (associated with prominence), phrasal edge tones, and boundary tones for the Intonational Phrase. Two Mandarin versions were proposed, C-ToBI (Li 2002) and Pan-Mandarin ToBI (Peng et al. 2005). Both versions included H% and L% as boundary tones and included register or pitch range specifications. Neither mentioned pitch accent – a local feature of a pitch contour, which is associated with the prominence of an utterance (Ladd 2008: 48). Jia (2009) attempted to propose an inventory of both pitch accents and boundary tones for Standard Mandarin intonation. However, the inventory of pitch accents did not tease apart lexical tones from intonation pitch accents. Zhang (2018b) examined polar question intonation and chanted call intonation in Tianjin Mandarin under the AM framework. Register differences and different types of boundary tones were proposed in the analysis.

Models such as the ones mentioned above depend heavily on how researchers transcribe the f0 contours into intonation categories. Another approach is to take a bottom-up approach to modelling intonation, which also addresses problems concerning intonation annotation.

### 8.5.2 Statistical modelling

While most traditional studies in prosody make use of linear modelling of acoustic parameters such as mean values of f0, f0 range, intensity, duration, among others, it is increasingly common to conduct dynamic analysis of the entire contour such as Growth Curve Analysis (GCA), General Additive Mixed Models (GAMM), and Functional Principal Component Analysis (fPCA).

Compared with linear modelling such as Linear Mixed Effect Models, GCA (e.g., Mirman et al. 2008) is able to account for non-linear data by bringing polynomial predictors into the modelling. This method has been applied in lexical tonal studies such as Chen et al. (2017), Li & Chen (2016), and Li et al. (2020).

GAMM (Wood 2017) is another method to compare two contours over time directly. It is more powerful than GCA in that not only are polynomial functions involved, but it can also combine different low-level functions as smoothing functions to better capture the data (Winter and Wieling 2016). For tone and intonation, GAMMs not only inform significant differences between tone/intonation categories but also illustrate which section of the f0 contours are different. Therefore, many recent prosody studies have used GAMMs, including Roy (2017), Zahner-Ritter et al. (2019; 2022), Deng et al. (2023), among many others.

FPCA also compares f0 contours directly. It is a dimension reduction algorithm which decomposes f0 contours into its main components. So, this method is especially suitable for studies based on AM theory, since it abstracts away from the phonetic details and presents the major contrasts across different categories. This method has also been applied in many prosody studies such as Asano and Gubian (2018), Chen and Boves (2018), Cheng et al. (2013), Lohfink et al. (2019), Bi & Chen (2022), and Zhang and Lahiri (in prep).

These models all rely on a continuous f0 contour rather than its discrete points and model the f0 changes over time. Increasingly, dynamic models have gained popularity. A few other possibilities include wavelet analysis (Suni et al. 2017), which is particularly successful in detecting features such as prosodic boundaries for speech technology applications; contour clustering (Kaland 2023), which is useful for detecting groups of prosodic contours, especially in exploratory studies. More recently, attempts have also been made to model and process f0 contours based on specific acoustic aspects of the signal. For example, Albert (2023) proposed an energy-based prosody analysis workflow and Iskarous et al. (2024) proposed an updated minimal dynamical model of intonation mainly based on velocity.

### 8.6 Discussion and conclusion

This chapter reviewed the findings on boundary phenomena in Mandarin Chinese and presented new data from naturally occurring speech to illustrate that even in a lexical tone language, where f0 is exploited to signal lexical tone contrasts, a relatively rich set of boundary phenomena based mainly on f0 events is still possible.

One issue that – on the one hand, complicates, and on the other, enriches – the discussion is that there are multiple ways of approaching intonation, as evident in the multitude of models proposed in the literature. In a tonal language, the situation is further complicated by the complex interactions of tone and intonation. Different prosodic models have, sometimes, drastically different views on almost every aspect. For instance, one topic briefly touched upon at the beginning of Section 8.4 is the differentiation between linguistic and paralinguistic functions. We further discuss it here since many additional boundary tone examples are conventionally regarded as belonging to paralinguistic prosody.  But how exactly are these potentially linguistic and paralinguistic functions conveyed in the boundary tones awaits further study. In the AM framework, Ladd (1996) made it very clear that one of the four major tenets of AM theory is that it is only concerned with linguistic function. PENTA, on the other hand, encodes both linguistic and emotional prosody in a parallel manner. The Kiel Intonation Model (Kohler 2006), among others, also does not distinguish paralinguistic and linguistic functions of prosody. This invites the question of whether or not it is necessary to separate the two. Furthermore, if it is indeed necessary, how do we draw the line between the two dimensions? A related topic is how to categorize the rich set of boundary phenomena. So far, we have followed the AM framework in describing the boundary tone forms based on peaks and valleys over an f0 contour. The assumed cognitive representation of these high and low boundary tones remains to be verified, and their relationship to intonational meanings is yet to be investigated.

To deepen our understanding of the tone-intonation interface, it is crucial to gather more data, particularly taking into consideration possible individual variations that have not been extensively explored in the literature (but see the individual variation of question-induced f0 raising/rising in Li & Chen 2025). These individual differences can provide valuable insights into the variability and adaptability of intonation patterns across speakers. Furthermore, while observing intonation phenomena in natural-occurring conversational speech is likely to yield more interesting examples than controlled studies, controlled and well-designed experiments are crucial for testing hypotheses and refining our understanding of the cognitive mechanisms underlying intonation representation and processing. An integrative approach that combines insights from naturalistic observations and designed hypothesis-testing experiments is crucial for advancing the field. In addition, given the dialectal variations in lexical tonal systems and the limited research on intonation across Mandarin varieties, more research is needed to explore dialect-specific intonation patterns (see some discussion in Chen 2022b). Knowledge as such is crucial for us to refine the typology of tone-intonation interactions. Last but not least, modeling will be essential for analyzing the data and gaining insights into these phenomena. In this chapter, we have focused on the production of intonation in Mandarin varieties, leaving perception, which is equally important, for future research. Ultimately, only with more data, clear goals, and robust models can we better understand how tone and intonation interact, both in production and beyond.